\pdfoutput=1
\documentclass[letterpaper, 10 pt, conference]{ieeeconf}  

\IEEEoverridecommandlockouts                              

\usepackage{cite}
\usepackage{amsmath,amssymb,amsfonts}
\usepackage{algorithmic}
\usepackage{graphicx}
\usepackage{textcomp}
\usepackage{xcolor}
\usepackage{booktabs}
\usepackage{multirow}
\usepackage{array}
\usepackage{tabularx}
\usepackage{stfloats}
\usepackage[normalem]{ulem} 
\usepackage{hyperref}  
\title{\LARGE \bf
Meta-Memory: Retrieving and Integrating Semantic-Spatial Memories for Robot Spatial Reasoning
}

\author{Yufan Mao, Hanjing Ye, Wenlong Dong, Chengjie Zhang, and Hong Zhang \\
        Southern University of Science and Technology, Shenzhen, China
       }

\begin{document}

\maketitle
\thispagestyle{empty}
\pagestyle{empty}

\begin{abstract}

Navigating complex environments requires robots to effectively store observations as memories and leverage them to answer human queries about spatial locations—a critical yet underexplored research challenge. While prior work has made progress in constructing robotic memory, few have addressed the principled mechanisms needed for efficient memory retrieval and integration. To bridge this gap, we propose Meta-Memory, a large language model (LLM)-driven agent that constructs a high-density memory representation of the environment. The key innovation of Meta-Memory lies in its capacity to retrieve and integrate relevant memories through joint reasoning over semantic and spatial modalities in response to natural language location queries, thereby empowering robots with robust and accurate spatial reasoning capabilities. To evaluate its performance, we introduce SpaceLocQA, a large-scale dataset encompassing diverse real-world spatial question-answering scenarios. Experimental results show that Meta-Memory significantly outperforms state-of-the-art methods on both the SpaceLocQA and the public NaVQA benchmarks. Furthermore, we successfully deployed Meta-Memory on real-world robotic platforms, demonstrating its practical utility in complex environments. Project page: \url{https://itsbaymax.github.io/meta-memory.github.io/}.

\end{abstract}

\section{INTRODUCTION}
A critical capability in robot navigation is the ability to understand and reason about human queries in order to identify and provide spatial locations as navigation destinations. This requires robots to construct a comprehensive memory from their sensory observations and to perform effective retrieval and reasoning over this memory. However, due to the complexity of real-world environments, building a complete and coherent visual representation from sequential observations remains a significant challenge. Even more demanding is the task of effectively utilizing the constructed memory for accurate spatial reasoning. In this work, we formalize these challenges as the Spatial Localization Question-Answering (SLQA) task—aiming to enable robots to build a holistic memory of large, complex environments and leverage it to answer diverse natural language queries about spatial relationships and locations. To the best of our knowledge, this is the first work to formally define the SLQA task.

Current approaches to robot memory construction typically rely on captions generated by Vision-Language Models (VLMs) \cite{team2023gemini,steiner2024paligemma,hurst2024gpt} or semantic embeddings extracted from foundational models \cite{radford2021learning,li2023blip}. These low-level representations are then used to build higher-level structures such as 3D scene graphs \cite{hughes2022hydra,gu2024conceptgraphs,werby2024hierarchical,xie2024embodied}, semantic maps \cite{shafiullah2022clip,huang2022visual,peng2023openscene}, or structured databases \cite{anwar2024remembr}. However, this paradigm faces several critical limitations.

First, significant information loss occurs during the encoding process. Even highly detailed captions and semantic embeddings inevitably fail to capture the full richness of raw sensory observations, leading to incomplete and potentially ambiguous memory representations.

More importantly, there is a notable lack of principled mechanisms for effective memory retrieval and integration—a core requirement for the SLQA task. Existing methods often treat memories as isolated fragments, lacking coherent organization or contextual linking. While some approaches, such as 3D scene graphs \cite{saxena2024grapheqa,deng2024opengraph} or topological maps \cite{yang20253d}, impose graph-based structures to encode spatial and semantic relationships, they still struggle with complex, multi-hop spatial reasoning. For instance, when a stranger on campus asks, ``Where's the nearest coffee shop southwest of here?", humans effortlessly retrieve the relevant memories, plan a route, and translate it into directions—an intuitive demonstration of the cognitive map \cite{tolman1948cognitive} in action. In contrast, current robotic systems remain far from achieving such flexible, context-aware spatial understanding.

\begin{figure}[t] 
  \centering
  \includegraphics[width=\linewidth]{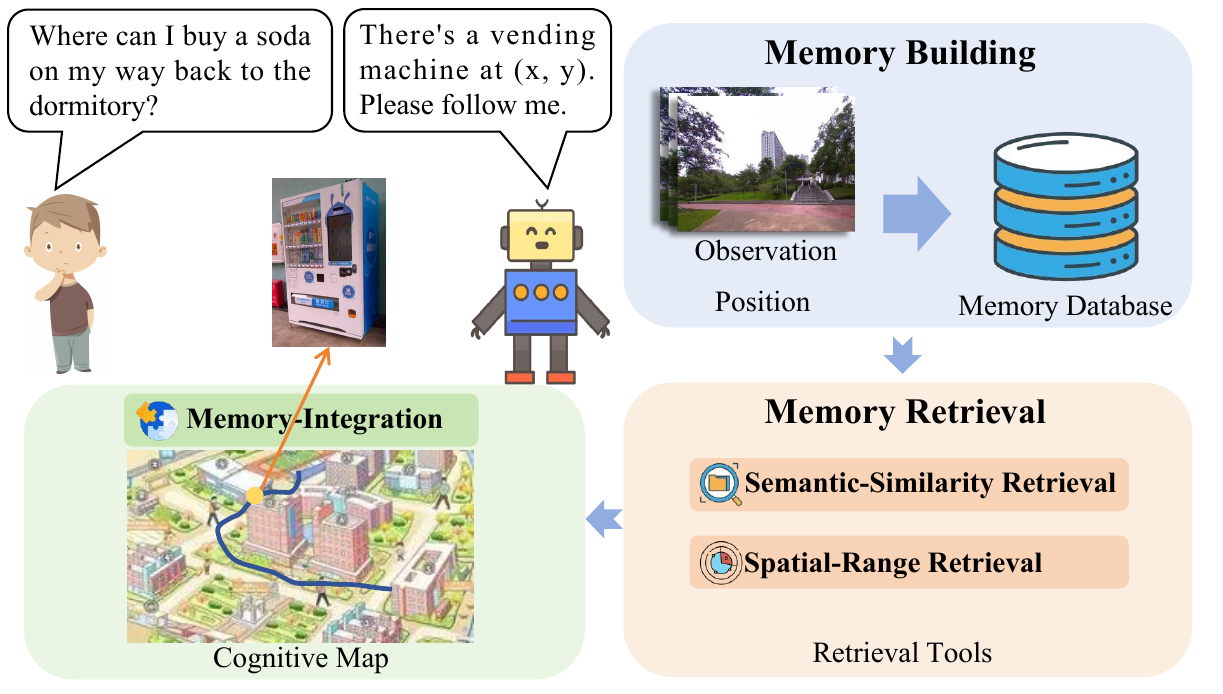}
  \vspace{-20pt}
  \caption{During its operation, Meta-Memory first constructs observations into memories. When it receives a location query from a user, Meta-Memory utilizes two distinct tools to retrieve memories along both semantic and spatial dimensions. Once sufficient information has been retrieved, Meta-Memory integrates the data to construct a cognitive map. Finally, it infers the target location based on this map.}
  \label{first_1}
\end{figure}

Recent work \cite{yang2025thinking} demonstrates that explicitly generating cognitive maps can significantly enhance the spatial reasoning capabilities of Multimodal Large Language Models (MLLMs). Inspired by this insight, we propose Meta-Memory, an LLM-based agent designed for the SLQA task. Our core idea is to enable the LLM agent to dynamically generate task-specific cognitive maps through comprehensive semantic and spatial retrieval over a rich, structured memory repository. To construct this memory system, we store the robot's raw sensory observations—comprising images and corresponding positions of the robot—as semantic-spatial memories. Building upon this foundation, the agent is equipped with two memory retrieval tools (semantic and spatial) and one memory integration tool. These tools facilitate thorough and fine-grained memory access, enabling the integration tool to synthesize a tailored cognitive map for each query. This structured representation strengthens the agent's spatial reasoning, leading to more accurate and contextually grounded responses. As illustrated in Fig.~\ref{first_1}, we present a complete reasoning pipeline of Meta-Memory, from perception and memory retrieval to cognitive map generation and response.

To evaluate our method, we conduct experiments on the spatial position questions from the NaVQA dataset\cite{anwar2024remembr}. To more rigorously assess the performance of various approaches on SLQA, we introduce SpaceLocQA, a new benchmark dataset. SpaceLocQA contains a broader and more diverse collection of real-world human spatial queries, enabling a comprehensive evaluation of a robot's ability to effectively construct, retrieve, and integrate its memories for accurate spatial inference.

The key contributions of this paper are:
\begin{itemize}
\item Meta-Memory, an LLM agent capable of comprehensively retrieving the constructed memories and effectively integrating the retrieved memories for spatial reasoning.
\item SpaceLocQA, a comprehensive spatial-localization dataset that encompasses diverse, realistic human location queries.
\item We deploy Meta-Memory on a physical robot and demonstrate its practical success at retrieving and integrating semantic-spatial memories in real-world environments.
\end{itemize}

\section{RELATED WORK}

\subsection{Semantic Memory and Retrieval}

Recent research\cite{yang20253d,deng2024opengraph,ge2025dynamicgsg} has made significant progress in semantic memory. Some studies\cite{ding2023pla,huang2022visual,shafiullah2022clip} have used foundational models trained on internet data to expand 2D semantics into 3D, creating semantic metric maps and enabling retrieval based on cosine similarity using text vectors within these maps. Additionally, 3D scene graphs\cite{gu2024conceptgraphs,ge2025dynamicgsg,werby2024hierarchical}, serving as a mental model for robots, construct high-level representations of entities and their relationships, allowing for reasoning over queries using LLMs. These 3D scene graphs provide robots with a semantically rich and structured representation of the environment, enhancing their ability to understand the surroundings and perform navigation and task planning. With the development of Retrieval-Augmented Generation (RAG)\cite{guo2024lightrag,edge2024local} in natural language processing, some studies\cite{dong2024rtagrasp,wang2025navrag} have extended it to the robotics domain. For example, specific studies\cite{xie2024embodied,anwar2024remembr} have used VLMs to generate captions from observations during robot navigation, constructing episodic memories and employing retrieval-based methods to better scale to long histories. In contrast to prior efforts that emphasize memory construction, our work investigates how to build a comprehensive semantic-spatial memory and, more importantly, how to effectively retrieve and integrate its contents to reason about arbitrary human spatial queries.

\subsection{Question Answering}

Video Question Answering (VQA)\cite{yang2025thinking} enables users to pose questions about video content using natural language, and the system generates accurate and relevant answers by understanding the visual information within the video. Embodied Question Answering (EQA)\cite{das2018embodied,majumdar2024openeqa} is an extension of VQA, requiring an agent to navigate and interact within a 3D environment to answer questions about the environment. Its core objective is to simulate how humans acquire knowledge while exploring unknown environments—specifically, the organic integration of language understanding, perception, decision-making, and reasoning. The work most similar to ours is NaVQA\cite{anwar2024remembr}, which leverages the robot's position and temporal information and focuses on generating navigation goals over longer historical context. Building upon the spatial-position subset of questions proposed in NaVQA, we introduce SpaceLocQA—a task that elevates spatial localization to the large scale. SpaceLocQA encompasses a broader and more realistic spectrum of spatial queries, demanding comprehensive, real-world spatial reasoning from embodied agents.

\subsection{Large Language Models for Navigation}

LLMs and VLMs have rapidly become key components for robot navigation. LLMs serve as high-level planners and commonsense knowledge bases, while VLMs typically provide rich perceptual features. In the exploration domain\cite{zhou2024navgpt,yang20253d}, for example, a VLM summarizes the current observation and an LLM decides the next move. Other works\cite{deng2024opengraph,gu2024conceptgraphs} integrate LLMs with 3D scene graphs to answer object-centric queries or to perform language-conditioned path planning. More recently, VLMs have been extended into Vision-Language-Action (VLA) models\cite{wang2025trackvla} that directly produce low-level actions while exhibiting strong generalization capabilities. Our work departs from prior art by building an LLM-based agent that employs a suite of specialized tools to tackle challenging SLQA task.

\begin{figure*}[ht] 
  \centering
  \includegraphics[width=\linewidth]{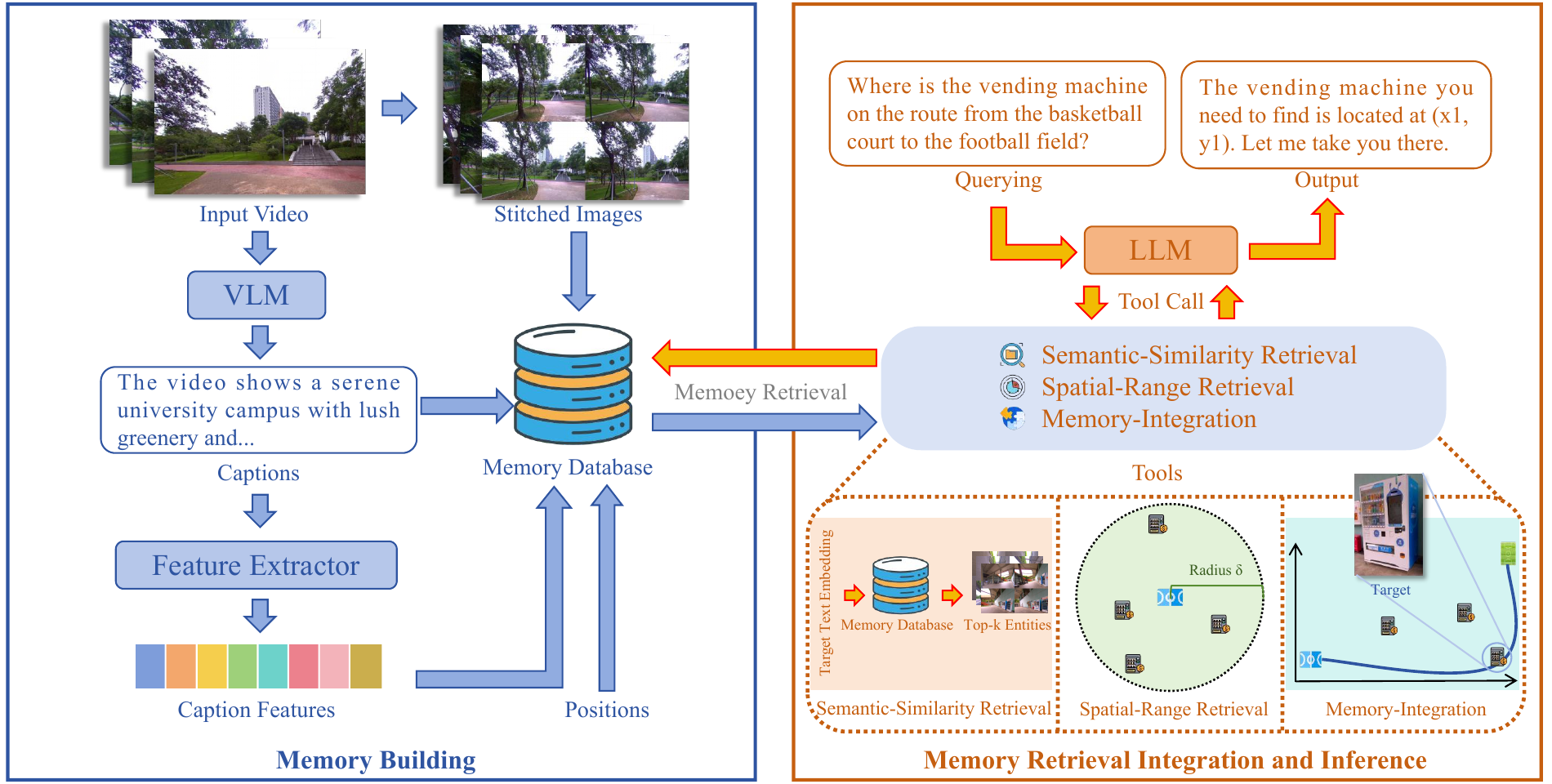}
  \vspace{-20pt}
  \caption{The Meta-Memory framework consists of two main modules: In the Memory Building phase, the Memory Database is constructed. In the Memory Retrieval Integration and Inference phase, the LLM interacts with the Memory Database using three tools to answer the query.}
  \label{method}
\end{figure*}

\section{METHOD}

To construct information-rich memories from the robot's observations and enable effective retrieval and integration for accurate localization in response to spatial queries, we propose Meta-Memory, a novel LLM-based framework. An overview is shown in Fig.~\ref{method}. Initially, we encode historical observations into semantic-spatial memories, which are enriched with detailed semantic and positional information (Sec. \ref{subsec:Memory Building}). To facilitate efficient memory retrieval based on a spatial location query, we introduce two memory retrieval tools: semantic-similarity retrieval (Sec. \ref{subsubsec:Semantic-Similarity Retrieval}) and spatial-range retrieval (Sec. \ref{subsubsec:Spatial-Range Retrieval}). These tools enable comprehensive memory retrieval across both semantic and spatial dimensions. Given the retrieved memories, we propose a memory-integration tool (Sec. \ref{subsubsec:Memory-Integration}). This tool integrates these memories into a coherent, task-specific cognitive map, thereby enhancing the agent's spatial reasoning capabilities. Finally, we detail the complete reasoning pipeline of the Meta-Memory system (Sec. \ref{subsubsec:Inference}).

\subsection{Memory Building}
\label{subsec:Memory Building}

In our memory-building phase, we define the robot's memory as a database. Unlike common approaches\cite{anwar2024remembr,xie2024embodied} that only use captions of observations as memories, we incorporate significantly richer visual information into the memory in an efficient manner. Concretely, for every t-second segment of the input video we (1) use a VLM to generate a caption and then embed that caption with the mxbai-embed-large-v1\footnote{https://www.mixedbread.com/blog/mxbai-embed-large-v1 } model; (2) within the same interval, extract four evenly spaced frames and concatenate them into a single image that acts as the robot's spatial-memory representation; and (3) store the caption, its embedding, the image, and the robot's position as a unified memory entry in the database. Both embeddings and positions are vectorized so that efficient vector-similarity search can be performed in subsequent stages.

\subsection{Memory Retrieval Integration and Inference}

Given a constructed memory and a spatial query, our Meta-Memory module retrieves and integrates relevant memory entries into a coherent cognitive map. First, it identifies memory entries that are both semantically similar and spatially proximate to the query. These entries are then synthesized and reassembled into a unified representation—namely, the cognitive map—enabling the agent to form a holistic understanding of its semantic-spatial knowledge and, ultimately, to infer the correct location.

\subsubsection{Semantic-Similarity Retrieval}\label{subsubsec:Semantic-Similarity Retrieval} 
We propose a coarse-to-fine memory retrieval mechanism that preserves the original sensory observations—specifically, the raw images—enabling an information-rich memory storage paradigm. Unlike prior approaches that rely solely on compressed representations such as VLM-generated captions, our method retains the full visual fidelity of the original inputs. This allows the agent to directly verify whether the image contains the queried object using a VLM. The retrieval pipeline proceeds as follows: (1) The LLM agent first extracts the textual description of the target object from the input query. (2) This text is encoded into a dense embedding vector E using the mxbai-embed-large-v1 model. (3) A Semantic-Similarity Retrieval (SSR) then searches the memory database DB using E, retrieving the top-k semantically similar memory entries, denoted $\text{ENTY}_k$. (4) GPT-4o is subsequently employed to examine the original images corresponding to $\text{ENTY}_k$, verifying whether they contain the object specified in the query. (5) Finally, a refined caption is generated and incorporated into the reasoning context to support downstream spatial inference. The core step of the above process can be abstracted into the following formula:
\begin{equation}
ENTY_k = SSR(DB \mid E) 
\end{equation} 

This strategy effectively mitigates the information loss inherent in caption-based or semantic-map-based memory representations. When only textual summaries (e.g., ``red cup", ``white cup", ``black cup") or semantic embeddings are stored, fine-grained visual attributes—such as shape, functional design, or subtle categorical distinctions—are often lost. For instance, if a query seeks a ``mug" that visually corresponds to a stored ``red cup", a purely caption-based retrieval system may fail due to semantic mismatch, despite strong visual similarity. Our approach overcomes this limitation by deferring detailed visual verification to a fine-grained reasoning stage. Moreover, our method follows a coarse-to-fine retrieval paradigm: rather than invoking a VLM to process all stored images directly—which would exceed the VLM's context length and incur prohibitive computational cost—we first retrieve a small set of candidate entries based on semantic relevance, and then apply the VLM only to these candidates. This design efficiently balances accuracy and scalability, effectively addressing the context and resource constraints of modern VLMs.

\subsubsection{Spatial-Range Retrieval}\label{subsubsec:Spatial-Range Retrieval} Suppose there are numerous objects in a kitchen scene. When generating a caption using a VLM, the model, similar to human attention, may overlook the presence of a particular cup. As a result, the caption will not contain information about the cup. After the LLM agent employs Semantic-Similarity Retrieval, if the retrieved memory does not include the target object, the LLM agent can select a position from the context that is most likely to be in the vicinity of the object specified in the query. This position, denoted as POS, is then used as the input for Spatial-Range Retrieval (SRR). SRR can autonomously determine the radius for retrieving memories from the database DB, resulting in a set of memories within the specified radius, denoted as $\text{MEM}_{\delta}$. For example, if the goal is to find a cup and the retrieved memory contains a water dispenser, the agent can retrieve memories within a 3-meter radius around the water dispenser. Similarly, if the goal is to find a restaurant, the agent might need to retrieve memories within a 50-meter radius around a supermarket.

Given that a larger retrieval radius yields a larger set of candidate memories $\text{MEM}_{\delta}$, directly applying the VLM to verify each memory would be computationally prohibitive. Instead, it first uses the LLM to filter the top-k most relevant memories using the captions in $\text{MEM}_{\delta}$, resulting in a subset $\text{MEM}_k$. Only then does the VLM confirm which original image contains the object specified in the query. Finally, a caption is generated and added to the context for further reasoning. The core process of this tool can be defined by the following formula:
\begin{equation}
\begin{aligned}
&MEM_\delta = SRR(DB \mid POS, \delta) \\ 
&MEM_k = LLM(MEM_\delta) \\ 
\end{aligned}
\end{equation}
where the parameter \text{$\delta$} represents the retrieval radius, determined by the LLM.

\subsubsection{Memory-Integration}\label{subsubsec:Memory-Integration} 
\begin{figure}[t] 
  \centering
  \includegraphics[width=\linewidth]{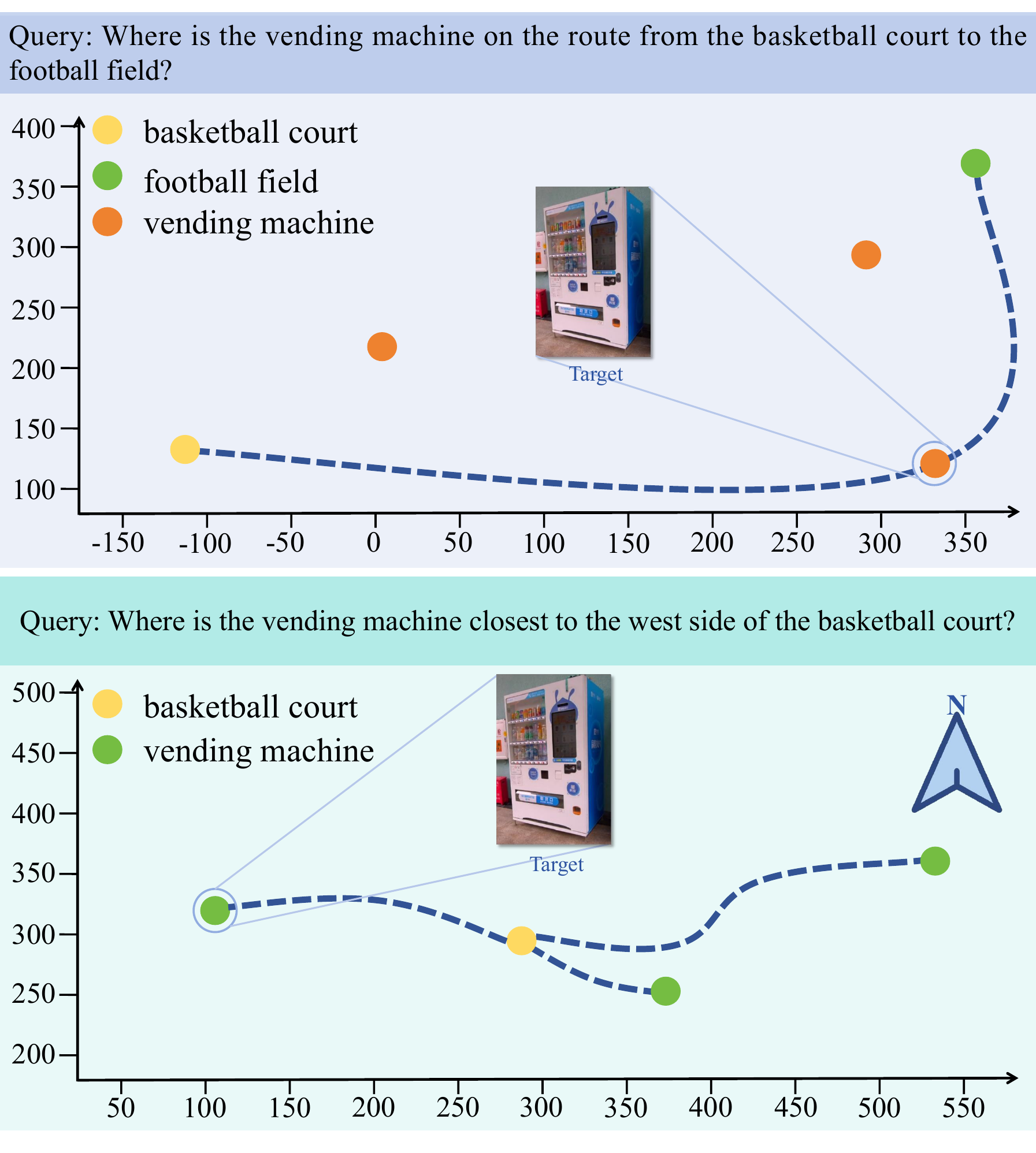}
  \vspace{-20pt}
  \caption{Based on the different queries and the input landmark positions, Memory-Integration can generate distinct cognitive maps.}
  \label{visual_prompt}
\end{figure}

This tool effectively integrates retrieved memories to construct a cognitive map for the robot, enabling structured representations with enhanced reasoning capabilities. The Memory-Integration (MI) autonomously identifies, based on the input query, the components required to build the cognitive map—such as waypoints, start landmarks, end landmarks, candidate landmarks, and directional indicators. As illustrated in Fig.~\ref{visual_prompt}, for a query like ``Where is the vending machine on the route from the basketball court to the football field?", the basketball court is designated as the start landmark, the football field as the end landmark, candidate vending machines serve as potential target landmarks, and the connecting waypoints are computed using Dijkstra's algorithm\cite{dijkstra2022note}. In contrast, for a query such as ``Where is the vending machine closest to the west side of the basketball court?", the basketball court acts as the start landmark, multiple vending machines are treated as end landmarks, and a directional indicator is incorporated to refine spatial constraints.

The procedure proceeds as follows: First, an LLM agent extracts key information from the contextual input, specifically identifying the positions (LM-POS) associated with each relevant landmark. Since the original memory database contains positional records of all observed entities, these can be directly utilized as navigable waypoints. A topological map is then constructed from these waypoints. Based on this topological structure, Dijkstra's algorithm is applied to compute the shortest paths (PATH) between LM-POS. With this information aggregated, the MI generates a query-specific cognitive map (COG-MAP), either through predefined rule-based programs or via a generative model. The resulting COG-MAP is subsequently fed into a VLM, which performs fine-grained inference to determine the precise location of the queried object. Finally, the inferred position is incorporated back into the context for future reference. The process of generating the COG-MAP can be represented by the following formula:
\begin{equation}
COG\text{-}MAP = MI(PATH, LM\text{-}POS, Q) 
\end{equation}

\begin{figure}[t] 
  \centering
  \includegraphics[width=\linewidth]{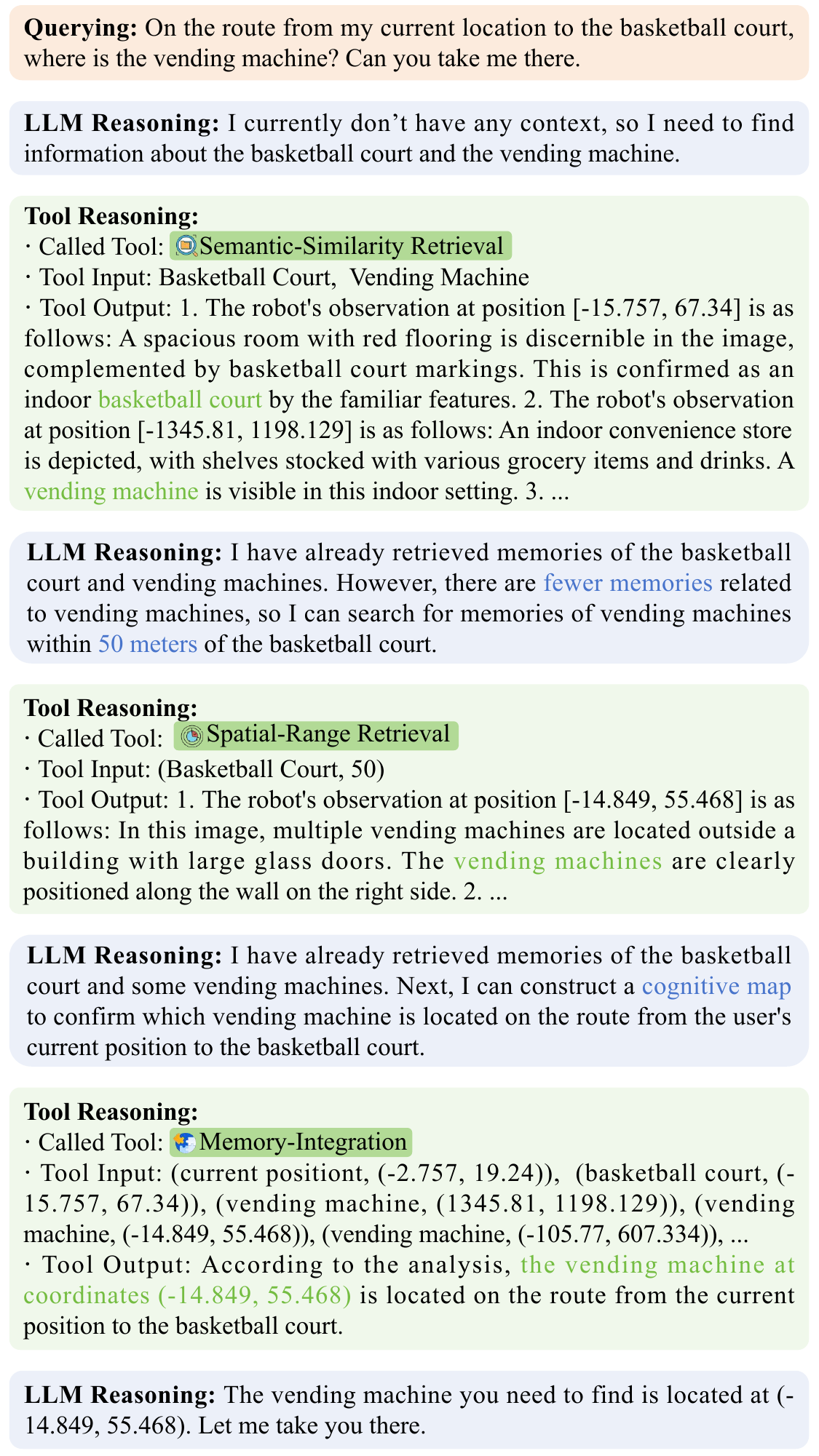}
  \vspace{-20pt}
  \caption{An example of inference by Meta-Memory. Starting from the Query, the LLM conducts reasoning to determine which tool to invoke and the inputs for that tool, proceeding with step-by-step reasoning until the answer is derived.}
  \label{inference}
\end{figure}

\subsubsection{Inference}\label{subsubsec:Inference} The inference process of Meta-Memory is similar to that of most works based on LLM agents. It begins with the input query, where the LLM decides on which tool to call and what inputs to provide to the tool. The output of the tool is then added to the context, after which the LLM continuously calls tools until an answer is generated or the maximum number of execution steps is reached. It is worth noting that, apart from Semantic-Similarity Retrieval, which is the most fundamental retrieval tool and must be called, the LLM agent can decide independently whether to use the other two tools. As illustrated in Fig.~\ref{inference}, an example is provided where all three tools are called. Meta-Memory is implemented using LangChain\footnote{https://www.langchain.com/}, with Milvus\footnote{https://milvus.io/} serving as the database and GPT-4o\cite{hurst2024gpt} as the LLM.

\section{EXPERIMENTAL SETUP}

\subsection{SpaceLocQA dataset}

To rigorously evaluate a robot's ability to construct and utilize memory for spatial localization question-answering, we introduce SpaceLocQA, a benchmark dataset comprising three distinct query categories: basic, local, and global. Basic queries test fundamental object retrieval (e.g., ``Where did I put my red cup?"), requiring precise recall of an object with specified attributes. Local queries examine short-range spatial understanding (e.g., ``Which room contains a refrigerator, a microwave, and a window?"), requiring the robot to integrate partial observations from multiple viewpoints into a coherent local representation. Global queries assess large-scale spatial reasoning (e.g., ``Where is the vending machine along the route from the basketball court to the soccer field?"), which necessitates long-horizon spatial perception and multi-step inference.

We captured six handheld video sequences, which total 68 minutes, across a university campus. Each sequence contributes 15 queries of each type, yielding a total of 270 queries. Accurate per-frame poses were obtained with Fast-LIO2\cite{xu2022fast} and subsequently transformed into the corresponding image coordinate system. To ensure high-quality annotations, three robotics specialists independently labeled the dataset, followed by a cross-validation process to verify consistency and correctness.

\subsection{Implementation details}

We conduct evaluations on two datasets: SpaceLocQA and NaVQA\cite{anwar2024remembr}. For SpaceLocQA—which features abundant scene text, including a significant proportion in Chinese—we employ Qwen2.5-VL-7B\cite{bai2025qwen2} as the video-captioning model, leveraging its strong multilingual text understanding capabilities. For NaVQA, we follow the approach of ReMEmbR\cite{anwar2024remembr} and employ VILA1.5-13B\cite{lin2024vila} for video captioning. Both VLMs operate on 3-second video clips, resulting in an effective frame rate of 2 FPS. GPT-4o serves as the LLM for all experiments.

\subsection{Metrics}

Meta-Memory and all baselines output (x, y) coordinates. On SpaceLocQA, a prediction is deemed successful if its Euclidean distance to the ground-truth location is less than 15 meters. For the spatial questions in NaVQA, we report the Euclidean distance between the predicted and ground-truth positions as the positional error, consistent with the original benchmark to reflect real-world navigation deviations. All experiments are conducted three times, and the average results are reported.

\subsection{Baselines}

We compare against three systems capable of constructing and querying semantic-spatial memory: ReMEmbR\cite{anwar2024remembr}, Embodied-RAG\cite{xie2024embodied}, and Humans.

ReMEmbR proposes a retrieval-augmented memory framework for embodied robots to support spatial question answering in long-horizon videos. It consists of two phases: memory construction and query processing, which aggregate temporal and spatial information to handle growing observation histories. During querying, an LLM-based agent retrieves relevant memory segments.

Embodied-RAG handles diverse queries across different environments, supporting reasoning about both specific objects and overall scene descriptions. Its memory is structured as a semantic forest, storing multi-level linguistic descriptions. During retrieval, an LLM hierarchically selects the most relevant node. For fair comparison, we equip each leaf node with the same VLM-generated captions used in our method.

Humans are advanced agents capable of constructing and retrieving semantic-spatial memory. We recruited three volunteers who had never visited the recorded environments. Each participant watched every video sequence three times. After viewing, they answered dataset questions by first identifying the most relevant memory and then selecting the corresponding image; the position associated with the selected image served as the final coordinate.

\section{RESULTS}

In this section, we primarily address two key questions: (1) Has our system bridged the gap between perception and memory representation by constructing a rich semantic-spatial memory? (2) Has our system effectively retrieved and integrated the constructed memory?

\begin{table*}[ht]
\centering
\caption{In the results comparing six scenarios on the SpaceLocQA dataset, our method surpassed the baseline in accuracy across all three types of queries, measured by success rate.}
\label{tab:performance1}
\begin{tabularx}{\textwidth}{>{\centering\arraybackslash}m{2cm} >{\centering\arraybackslash}m{2cm} *{6}{>{\centering\arraybackslash}X} >{\centering\arraybackslash}X}
\toprule
\textbf{Category} & \textbf{Method} & \textbf{0} & \textbf{1} & \textbf{2} & \textbf{3} & \textbf{4} & \textbf{5} & \textbf{Average$\uparrow$} \\ \midrule
\multirow{4}{*}{Basic} & ReMEmbR & 46.7 & 53.3 & 82.2 & 71.1 & 40.0 & 57.8 & 58.5 \\ 
 & Embodied-RAG & 42.2 & 37.8 & 71.1 & 62.2 & 35.6 & 68.9 & 53.0 \\ 
 & Human & 31.1 & 55.6 & 62.2 & 48.9 & 55.6 & 71.1 & 54.1 \\ 
 & Meta-Memory & 53.3 & 57.8 & 84.4 & 73.3 & 57.8 & 80.0 & \textbf{67.8(+9.3)} \\ \midrule
\multirow{4}{*}{Local} & ReMEmbR & 68.9 & 60.0 & 51.1 & 66.7 & 44.4 & 55.6 & 57.8 \\ 
 & Embodied-RAG & 62.2 & 48.9 & 57.8 & 53.3 & 46.7 & 62.2 & 55.2 \\ 
 & Human & 28.9 & 51.1 & 55.6 & 57.8 & 31.1 & 44.4 & 44.8 \\ 
 & Meta-Memory & 66.7 & 60.0 & 53.3 & 64.4 & 60.0 & 66.7 & \textbf{61.8(+4.0)} \\ \midrule
\multirow{4}{*}{Global} & ReMEmbR & 51.1 & 44.4 & 60.0 & 37.8 & 51.1 & 33.3 & 46.3 \\ 
 & Embodied-RAG & 37.8 & 42.2 & 40.0 & 20.0 & 40.0 & 44.4 & 37.4 \\ 
 & Human & 35.6 & 40.0 & 48.9 & 57.8 & 48.9 & 64.4 & 49.3 \\ 
 & Meta-Memory & 60.0 & 62.2 & 82.2 & 55.6 & 60.0 & 53.3 & \textbf{62.2(+12.9)} \\ \bottomrule
\end{tabularx}
\end{table*}

\subsection{Experimental results}

Table \ref{tab:performance1} presents the evaluation results of Meta-Memory and the baseline methods on the SpaceLocQA dataset. As can be clearly seen in the table, Meta-Memory outperforms the baseline methods across all three types of queries. Particularly for global queries, Meta-Memory demonstrates a significant advantage. This improvement is likely attributed to Meta-Memory's ability to integrate retrieved memories into a structured cognitive map, enabling relational and topological reasoning beyond simple retrieval. In basic and local queries, the baseline methods suffer from lower accuracy due to the loss of some details in the constructed memories and the inability to effectively retrieve these memories.

The Human baseline is particularly interesting. Due to limited human attention, not all elements in a scene can be stored as memories. Moreover, humans' tendency to forget information is a disadvantage compared to robots. As a result, the human baseline performs poorly on the SpaceLocQA dataset. An interesting observation is that in global queries, humans, similar to Meta-Memory, have the ability to integrate memories to construct a cognitive map. Therefore, the human baseline outperforms the other two baseline methods in this type of query. However, in local queries, which require strong detailed memory capabilities, local spatial modeling abilities, and long-horizon memory capabilities, the human baseline is the worst among the compared methods.

\begin{figure*}[t] 
  \centering
  \includegraphics[width=\linewidth]{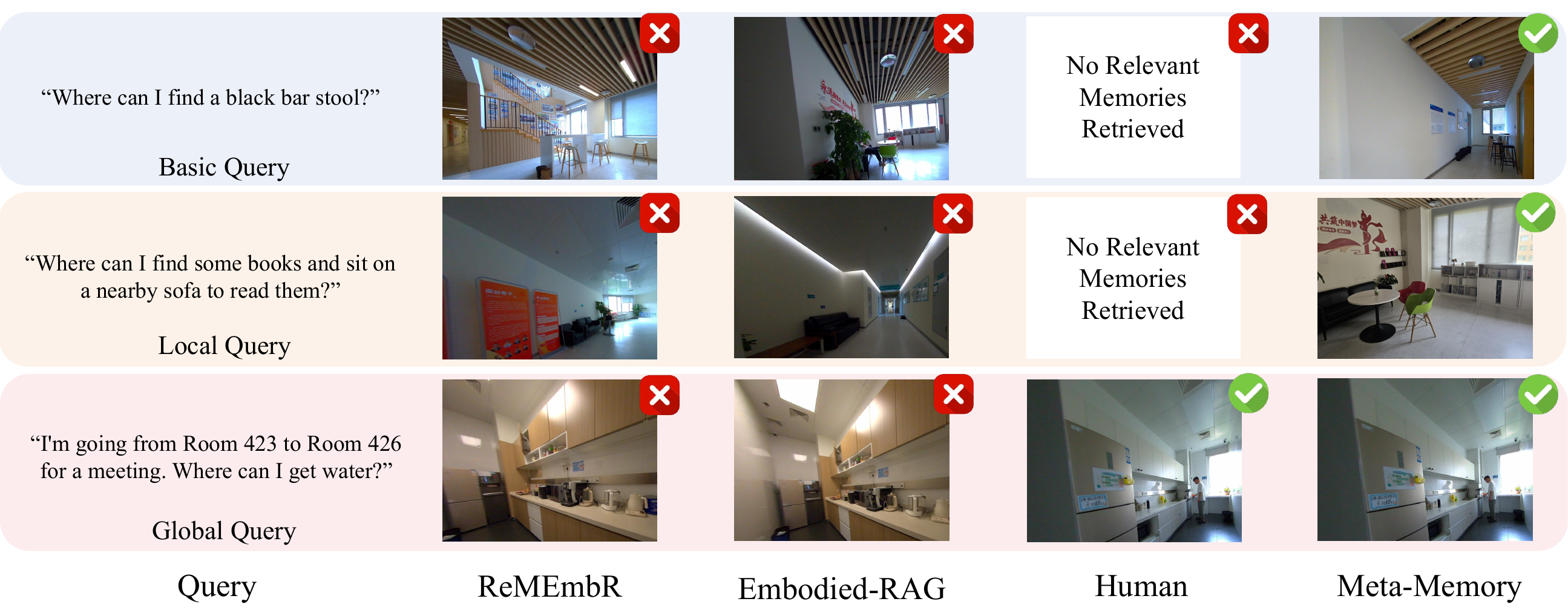}
  \vspace{-20pt}
  \caption{Examples of the three types of queries in SpaceLocQA. We present the images corresponding to the (x, y) coordinates outputted by the system. In the examples of human output results, ``No Relevant Memories Retrieved" indicates that the human volunteer failed to retrieve relevant memories.}
  \label{outcome}
\end{figure*}

Fig.~\ref{outcome} intuitively shows the comparison results between Meta-Memory and other methods. When answering a basic query, both ReMEmbR and Embodied-RAG retrieve memories that lack fine-grained details. Although they retrieve a chair, it does not match the description of a ``black bar stool" in the query. In the case of a local query, ReMEmbR can only retrieve discrete memories and lacks the ability to integrate these memories. Therefore, it falls short in perceiving the local surrounding environment. Embodied-RAG, despite structuring its memories into a semantic forest, fails in this case because it did not capture the ``book" during memory construction, highlighting the importance of constructing a high-density memory. In the global query example, it is evident that both ReMEmbR and Embodied-RAG lack the ability to effectively integrate memories. Neither can identify the water cooler located on the path from Room 423 to Room 426. In contrast, both humans and Meta-Memory can easily locate the correct water cooler.

Table \ref{tab:performance2} presents the positional-error results on NaVQA (excluding the Human baseline results). By directly measuring the Euclidean distance between predicted and ground-truth coordinates, Meta-Memory again exhibits a substantial advantage, underscoring the superiority of our approach.

\subsection{Ablations}

Table \ref{tab:performance3} isolates the contribution of each component. For basic queries, removing the raw image memory (i.e., not storing observed images) results in a sharp drop in accuracy, suggesting that fine-grained visual details preserved in raw images are critical for answering basic queries. For global queries, disabling the Memory-Integration tool (i.e., preventing cognitive-map construction) reduces performance to nearly baseline levels, highlighting its critical role in enabling long-range spatial reasoning. In addition, disabling Spatial-Range Retrieval also lowers average accuracy, indicating that retrieval along the spatial dimension is likewise necessary.

\begin{table}[t]
\centering
\caption{On the NaVQA dataset, Meta-Memory outperformed the other two methods with a lower mean Euclidean error, indicating more accurate predictions.}
\label{tab:performance2}
\begin{tabularx}{\linewidth}{>{\centering\arraybackslash}p{1.9cm} *{7}{>{\centering\arraybackslash}p{0.34cm}} >{\centering\arraybackslash}p{0.39cm}}
\toprule
\textbf{Method} & \textbf{0} & \textbf{3} & \textbf{4} & \textbf{6} & \textbf{16} & \textbf{21} & \textbf{22} & \textbf{Mean$\downarrow$} \\ \midrule
ReMEmbR & 27.5 & 13.3 & 23.5 & 57.9 & 29.6 & 33.2 & 14.5 & 28.5 \\ 
Embodied-RAG & 31.8 & 9.4 & 33.5 & 33.4 & 43.4 & 37.1 & 29.4 & 31.1 \\ 
Meta-Memory & 17.1 & 5.6 & 18.7 & 37.5 & 36.9 & 22.8 & 13.4 & \textbf{21.7} \\ 
\bottomrule
\end{tabularx}
\end{table}

\begin{table}[t]
\centering
\caption{The comparison results between Meta-Memory with various modules removed and the original Meta-Memory.}
\label{tab:performance3}
\begin{tabularx}{\linewidth}{l *{4}{>{\centering\arraybackslash}X}}
\toprule
\textbf{Configuration} & \textbf{Basic} & \textbf{Local} & \textbf{Global} & \textbf{Average$\uparrow$} \\ \midrule
w/o Raw Image Memory & 60.7 & 58.5 & 56.3 & 58.5 \\ 
w/o Spatial-Range Retrieval & 65.2 & 60.7 & 59.3 & 61.7 \\ 
w/o Memory-Integration & 67.8 & 61.5 & 49.6 & 59.6 \\
Meta-Memory & \textbf{67.8} & \textbf{61.8} & \textbf{62.2} & \textbf{63.9} \\ 
\bottomrule
\end{tabularx}
\end{table}

\begin{figure}[t] 
  \centering
  \includegraphics[width=\linewidth]{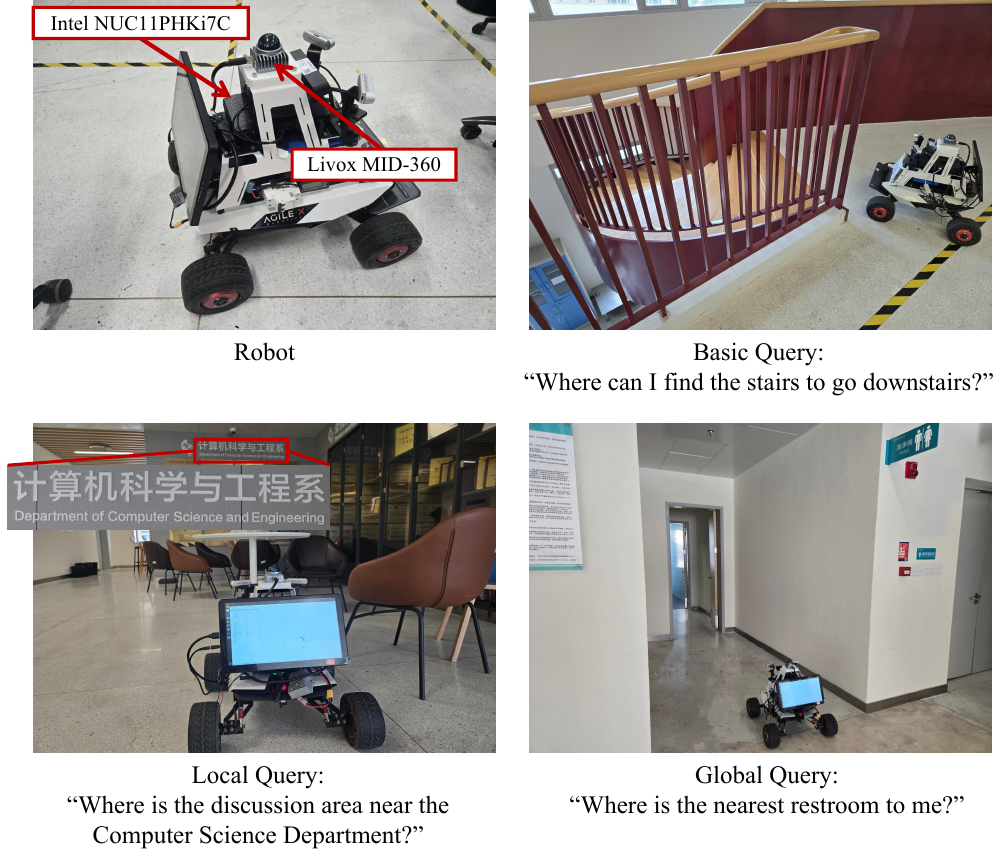}
  \vspace{-20pt}
  \caption{We loaded the offline-constructed memory onto the AgileX Scout Mini robot and then ran Meta-Memory to handle various types of queries. The robot stored a memory containing a large amount of detailed information and successfully navigated to the destination specified in the query by utilizing this memory.}
  \label{deployment}
\end{figure}

\subsection{Real-world Deployment}
As illustrated in Fig.~\ref{deployment}, we deployed Meta-Memory on the AgileX Scout Mini robot\footnote{https://global.agilex.ai/products/scout-mini} to demonstrate its ability to successfully navigate to target locations in real-world environments. Specifically, memory construction was performed offline and subsequently transferred to the robot. During inference, we employed GPT-4o as the LLM and deployed the mxbai-embed-large-v1 model to compute the embedding vector of the input query. Upon receiving a natural language query, Meta-Memory outputs a 2D coordinate representing the estimated target location. Since the original memory stored the position corresponding to each observation, we constructed a topological map from it. Based on this topological map, we used Dijkstra's algorithm to calculate the shortest path and generate multiple waypoints. The robot acquired point clouds using Livox-Mid-360\footnote{https://www.livoxtech.com/mid-360} and ran Fast-LIO\cite{xu2021fast} to obtain its pose. Based on the acquired pose, we used NeuPan\cite{han2025neupan} to output actions to navigate to the goal. It is worth noting that all of the above, except for the remote invocation of GPT-4o, can be run on an Intel NUC11PHKi7C (Intel Core i7, NVIDIA RTX2060, 6GB VRAM).

\section{CONCLUSION}

This work introduces Meta-Memory, a novel framework for constructing and leveraging semantic-spatial memory in robotic navigation. It preserves raw sensory observations as the foundation of memory. Built upon this, three specialized modules enable an LLM agent to dynamically retrieve, contextualize, and synthesize relevant memories as needed. In summary, for spatial localization question-answering task, Meta-Memory offers a state-of-the-art approach to memory management, enhancing the reasoning, adaptability, and capabilities of intelligent agents.

\textbf{Limitations and Future Work.} Meta-Memory has several limitations. Most notably, LLMs remain prone to hallucinations, which can lead to erroneous reasoning. Conceptually, the current system is designed for static environments and lacks the capability to handle dynamic scenes. Future work will explore more effective memory retrieval methods beyond text embedding models. Additionally, we plan to enhance adaptability to dynamic environments by updating the memory through retrieving the frame with the maximum overlap relative to the current frame at each step of the robot's memory construction process.

\addtolength{\textheight}{-6.5cm}   








\bibliographystyle{IEEEtran} 
\bibliography{IEEEabrv,root}

\end{document}